\documentclass[conference]{IEEEtran}
\IEEEoverridecommandlockouts
\usepackage{cite}
\usepackage{amsmath,amssymb,amsfonts}
\usepackage{algorithmic}
\usepackage{graphicx}
\usepackage{makecell}
\usepackage{textcomp}
\usepackage{xcolor}
\usepackage{graphicx}
\usepackage{booktabs}
\usepackage{threeparttable}
\usepackage{multirow}
\usepackage{diagbox}
\usepackage{float}
\def\BibTeX{{\rm B\kern-.05em{\sc i\kern-.025em b}\kern-.08em
    T\kern-.1667em\lower.7ex\hbox{E}\kern-.125emX}}
\begin{document}

\title{Multi-stage feature decorrelation constraints for improving CNN classification performance\\
}
\author{\IEEEauthorblockN{1\textsuperscript{st} Qiuyu Zhu}
\IEEEauthorblockA{\textit{School of Communication \& Information Engineering} \\
\textit{Shanghai University}\\
Shanghai, China \\
zhuqiuyu@staff.shu.edu.cn}\\

\IEEEauthorblockN{3\textsuperscript{rd} Xuewen Zu}
\IEEEauthorblockA{\textit{School of Communication \& Information Engineering} \\
\textit{Shanghai University}\\
Shanghai, China \\xuewenzu@shu.edu.cn
}
\and
\IEEEauthorblockN{2\textsuperscript{nd} Hao Wang}
\IEEEauthorblockA{\textit{School of Communication \& Information Engineering} \\
\textit{Shanghai University}\\
Shanghai, China \\wanghao1998@shu.edu.cn
}\\

\IEEEauthorblockN{4\textsuperscript{th} Chengfei Liu}
\IEEEauthorblockA{\textit{School of Communication \& Information Engineering} \\
\textit{Shanghai University}\\
Shanghai, China \\
liuchengfei@shu.edu.cn}
}

\maketitle

\begin{abstract}
For the convolutional neural network (CNN) used for pattern classification, the training loss function is usually applied to the final output of the network, except for some regularization constraints on the network parameters. However, with the increasing of the number of network layers, the influence of the loss function on the network front layers gradually decreases, and the network parameters tend to fall into local optimization. At the same time, it is found that the trained network has significant information redundancy at all stages of features, which reduces the effectiveness of feature mapping at all stages and is not conducive to the change of the subsequent parameters of the network in the direction of optimality. Therefore, it is possible to obtain a more optimized solution of the network and further improve the classification accuracy of the network by designing a loss function for restraining the front stage features and eliminating the information redundancy of the front stage features .For CNN, this article proposes a multi-stage feature decorrelation loss (MFD Loss), which refines effective features and eliminates information redundancy by constraining the correlation of features at all stages. Considering that there are many layers in CNN, through experimental comparison and analysis, MFD Loss acts on multiple front layers of CNN, constrains the output features of each layer and each channel, and performs supervision training jointly with classification loss function during network training. Compared with the single Softmax Loss supervised learning, the experiments on several commonly used datasets on several typical CNNs prove that the classification performance of Softmax Loss+MFD Loss is significantly better. Meanwhile, the comparison experiments before and after the combination of MFD Loss and some other typical loss functions verify its good universality. The code for the paper can be found at https://github.com/lovelyemperor/MFD.
\end{abstract}

\begin{IEEEkeywords}
Feature decorrelation, CNN, Pattern classification,  Loss function , Feature redundancy
\end{IEEEkeywords}

\section{Introduction}
Convolutional neural network (CNN) is a deep learning model that has attracted wide attention. It can be used to solve various tasks such as image classification~\cite{b1}, face recognition~\cite{b2}, object detection~\cite{b2,b3}, and has very good performance. Loss function is an important part in the training process of CNN, which determines the accuracy of network training. Commonly used loss functions include Softmax Loss, MSE Loss, Focal Loss~\cite{b3}, etc. The classical Softmax Loss is widely used in image classification because of its fast convergence speed and good performance.

In the application of pattern recognition, CNN structure generally includes convolutional layers and fully connected layers, where convolutional layers are used to extract the features of input information, and fully connected layers are used for classification. The loss functions mentioned above restrict the output of the last fully connected layer or the last hidden layer, so the impact of these loss functions on front layers is gradually attenuated, and the adjustment of network parameters is not optimal. In the previous network training methods, there are some regularization methods~\cite{b4,b5} for the constraint of the network parameters, but as far as we know, the constraint methods for the features of the network front layers have not been seen, which makes it difficult to optimize the features of the front layers.

On the other hand, there is significant redundant information in the feature map of the well-trained neural network~\cite{b6}. For example, Fig. 1 presents some feature maps of an input image generated by ResNet50~\cite{b7} (after Softmax Loss supervised learning), and there exist many similar feature maps. Taking advantage of the characteristic, ~\cite{b6} proposes a new ghost module, which uses a series of low-cost linear transformations to generate many ghost feature maps based on a set of internal feature maps to improve the network operation speed. ~\cite{b8,b9} propose to prune the unimportant connections in the neural network. Channel pruning~\cite{b10,b11} further targets on removing the useless channels for easier to accelerate in practice. These are all optimized from the aspect of improving network speed, but there is no relevant research on how to use the feature correlation to improve network classification accuracy.

\begin{figure}[htbp]
\centering
\includegraphics[width=0.35\textwidth]{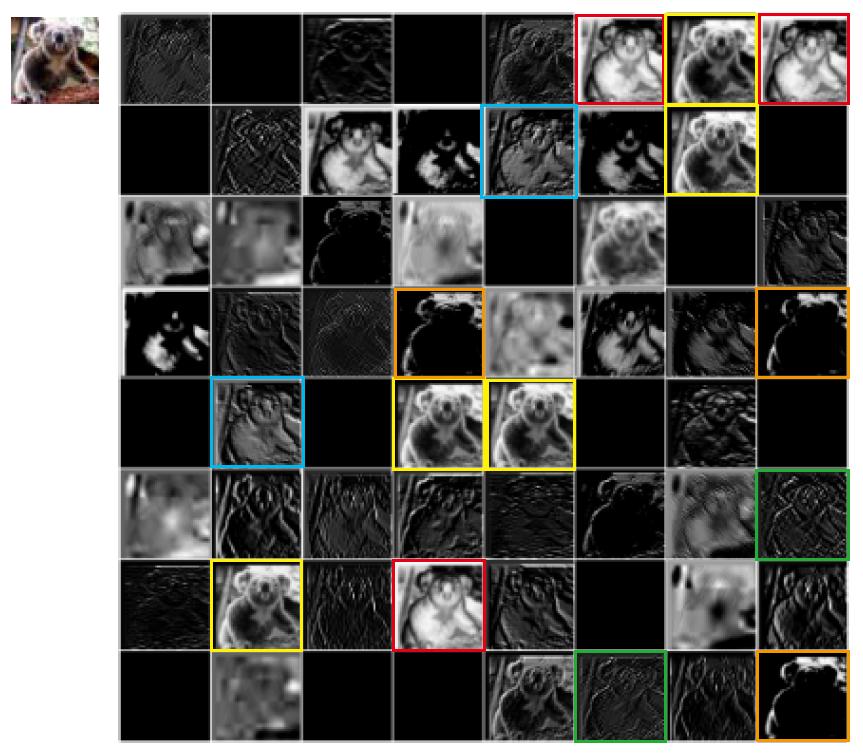}
\caption{Visualization of some feature maps generated after the first convolution group in ResNet50, where similar feature maps are annotated with boxes of the same color.}
\label{fig1}
\end{figure}

\begin{figure}[htbp]
\centering
\includegraphics[width=0.45\textwidth]{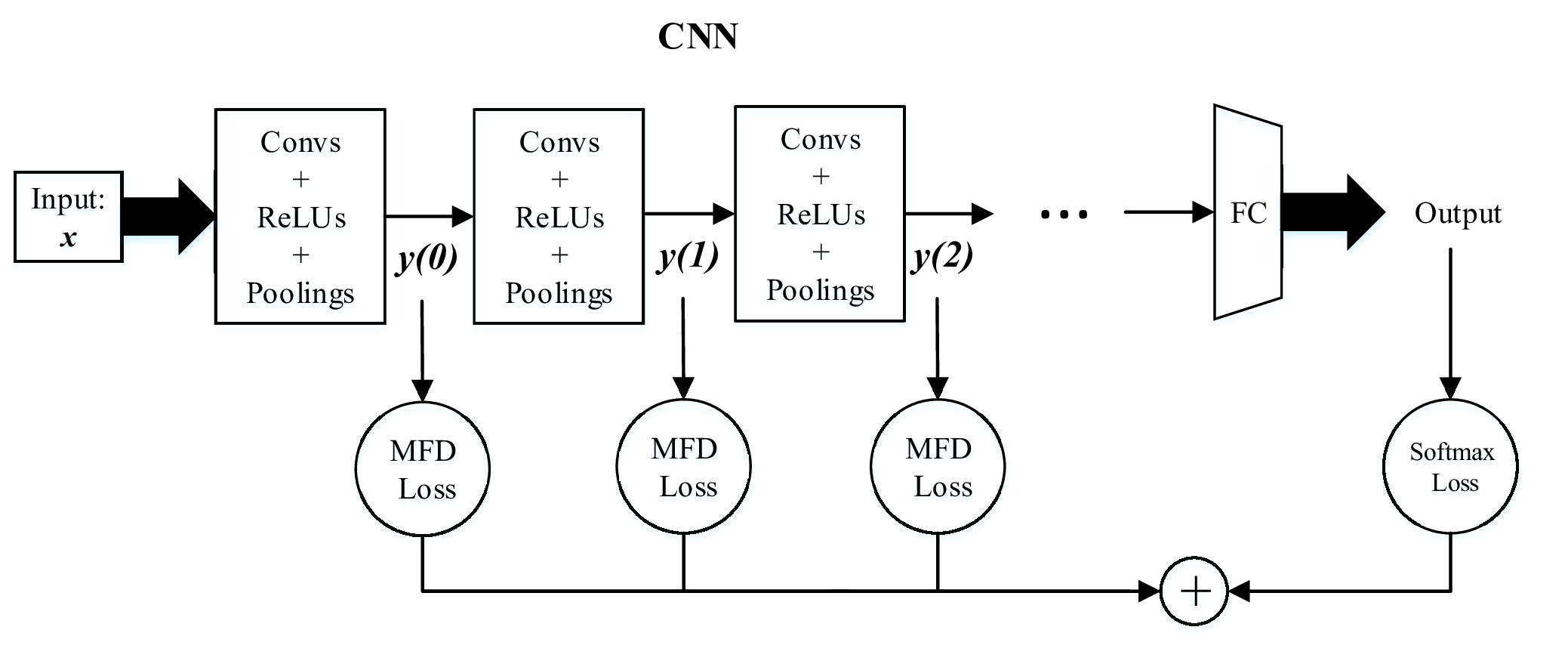}
\caption{MFD Loss and Softmax Loss in CNN.}
\label{fig2}
\end{figure}

In this article, a new loss function for the front layer features, multi-stage feature decorrelation loss (MFD Loss), is proposed to effectively reduce the redundancy of the front layer features in neural networks and refine effective features. Specifically, the features’ correlation of multiple front layers in CNN is calculated. In the training process, the neural network is trained under the joint supervision of Softmax Loss and MFD Loss. Intuitively, Softmax Loss forces the training direction to be correct, and MFD Loss reduces the redundancy of multiple front layers features, makes the features more effective, so that the parameters of the whole network are updated in the global optimal direction, thus improves the classification performance.

The location of MFD Loss and the entire network structure are shown in Fig. 2. The details of the method are given in Section 3.

Our main contributions are as follows:
\begin{enumerate}[]
\item A new loss function (MFD Loss) is proposed to maximize the feature validity of multiple front layers. As far as we know, this is the first time to constrain multiple front layers features in the neural network to help supervise the learning of CNN. The experimental results show that adding MFD Loss to classification loss function can achieve better classification performance.
\item For the classification task, extensive experiments on several commonly used datasets have carried out, and verified the excellent performance of our method on several typical CNN structures.
\item For the face recognition task, we compare the classification performance of several typical loss functions dedicated to face recognition before and after combining with MFD Loss, and demonstrate the good generalizability of our method.
\end{enumerate}

\section{Related Work}

In the field of supervised learning, the loss function is used to measure the similarity between the predicted value of the model and the true value. The smaller the loss function, the better the robustness of the model. The loss function guides the learning of the model, and assists in updating the parameters of the model during the training phase.

There are various loss functions in the supervised classification learning of neural networks. The most common loss function is Softmax Loss, which is a combination of Softmax and Cross-entropy Loss. The formula is as follows:
\begin{equation}
L_{Softmax}=\frac{1}{N}\sum_{i} -ln{\frac{e^{z_{y_i}}}{\sum_{j}e^{z_j}}}
\end{equation}
where $N$ is the number of samples, $z_{y_i}$is the output value of the last fully connected layer of the correct class $y_i$, and $z_j$ is the output value of the last fully connected layer of class $j$. Because $z_{y_i}=W^{T}_{y_i}x_i$, that is, $z_{y_i}=||W^{T}_{y_i}||||x_i||cos(\theta_{y_i})$, $W^{T}_{y_i}$ is the corresponding weight in the fully connected layer, and $x_i$ is the input characteristic of the $i$-th sample. Therefore, the Softmax loss function becomes the following formula:
\begin{equation}
L_{Softmax}=\frac{1}{N}\sum_{i} -ln{\frac{e^{||W^{T}_{y_i}||||x_i||cos(\theta_{y_i})}}{\sum_{j}e^{||W^{T}_j||||x_i||cos(\theta_j)}}}
\end{equation}

On the basis of Softmax, L-Softmax ~\cite{b12}, AM-Softmax ~\cite{b13}, and Center Loss ~\cite{b14} have improved it, but mainly for specific applications such as face recognition, and are not universally applicable.

In the field of unsupervised learning, Barlow Twins~\cite{b15} self-supervised learning algorithm proposes an innovative loss function, which adds a decorrelation mechanism to eliminate the correlation of the network output features and maximize the variability of representation learning. Barlow Twins Loss can be expressed as:
\begin{equation}
L_{BT}=\sum_{i}(1-C_{ii})^2+\lambda\sum_{i}\sum_{j \neq i}C^2_{ij}
\end{equation}
where $\lambda$ is a weighting coefficient used to weigh two items in the loss function, and $C$ is the cross-correlation matrix of the output features of the samples and their enhanced samples under the same batch of two identical networks. The addition of the latter redundancy reduction item in the loss function reduces the redundancy between the network output features, so that the output features contain non-redundant information of the sample, and achieves a good feature representation effect.

All the above loss functions act on the output features of the last fully connected layer of the network. The impact on the front layer features is achieved through the back-propagation algorithm, so the further forward the network layer is, the weaker the influence by the loss function, and the easier it is for the network to fall into a local optimum solution.

\section{Method}

\subsection{Correlation between Features}\label{AA1}
In general, the hidden features of all layers in a fully trained neural network have some correlation, which indicates the existence of information redundancy among these hidden features. We believe that the target of subsequent network layers’ parameters adjustment is not clear enough due to the information redundant feature inputs, so it has a negative impact on the network classification performance. We think that adding constraints between features can refine effective features, remove information redundancy, and ensure efficient parameter tuning in the subsequent network layer, making subsequent features more beneficial for classification. In this case, in order to improve the effectiveness of the features, we suggest that it is necessary to constrain the correlation between the features.

\subsection{Pearson Correlation Coefficient}
In statistics, Pearson correlation coefficient is used to measure the degree of correlation between two variables, and its value is between $-1$ and $1$. The value of $1$ indicates a completely positive correlation between the two random variables. The value of $-1$ indicates a completely negative correlation between the two random variables. The value of $0$ indicates that there is no linear correlation between the two random variables. The calculation formula is as follows:
\begin{equation}
r=\frac{\sum_{i=1}^n(X_i-\bar{X})(Y_i-\bar{Y})}{\sqrt{\sum_{i=1}^n(X_i-\bar{X})^2}\sqrt{\sum_{i=1}^n(Y_i-\bar{Y})^2}}
\end{equation}
where $r$ represents the value of Pearson correlation coefficient, $n$ is the number of samples, $X_i$ and $Y_i$ are the $i$-point observation values corresponding to variables $X$ and $Y$, $\bar{X}$ and $\bar{Y}$ is the average value of $X$ and $Y$ samples.
\subsection{Multi-stage Feature Decorrelation Loss}

Barlow Twins Loss~\cite{b15} adds a decorrelation mechanism to reduce the redundancy between network output features, so that the output features contain non-redundant information of samples. Barlow Twins Loss uses the cross-correlation matrix and punishes the non-diagonal terms of the computed cross-correlation matrix. The decorrelation loss proposed in this subsection uses Pearson correlation coefficient in subsection 3.2 to calculate the correlation degree between features, to obtain the correlation coefficient matrix, and penalizes the non-diagonal terms of the matrix to constrain the correlation between features.

Features at each layer in CNN contain batch size, dimension (number of channels), two-dimensional feature map and so on, so the multi-stage feature decorrelation loss function can be expressed as:
\begin{equation}
L_{MFD}=\frac{\sum_{i}^d\sum_{j \neq i}^d F^2_{ij}}{d(d-1)}
\end{equation}
\begin{equation}
F_{IJ}=\frac{\sum_{k=1}^b\langle I_k-\bar{I}, J_k-\bar{J}\rangle}{\sqrt{\sum_{k=1}^b(I_k-\bar{I})^2}\sqrt{\sum_{k=1}^b(J_k-\bar{J})^2}}
\end{equation}
where $d$ represents the number of feature dimension, $F$ represents the Pearson correlation coefficient matrix formed between the features. The elements in the $i$-th row and the $j$-th column of the matrix are the correlation coefficients of the $i$-th and $j$-th features. $b$ is the batch size. $I_k$ and $J_k$ represent the observed values of the $k$-th samples of the $I$-th and $J$-th features, which are a two-dimensional feature map with different resolutions at each stage. $\bar{I}$ and $\bar{J}$ represent the average values of the $I$-th and $J$-th features in each batch. $\langle A,B \rangle$ represents the inner product of two matrices $A$ and $B$, which is the sum of the corresponding elements multiplied, resulting in a scalar.

Using the typical network structure ResNet50~\cite{b7}, the convolutional part of ResNet50 can be divided into $5$ stages, which can be calculated separately for the output of each stage. As shown in Fig. 3, the structure of Stage $0$ is comparatively simple, which is the preprocessing of input data, and the structures of the last four stages are similar. After the CIFAR100 dataset is fully trained by Softmax Loss, we calculated the average values of correlation coefficients between features after Stage $0$ to Stage $4$. Due to the range of correlation coefficients between $-1$ and $1$, in order to avoid mutual cancellation when averaging correlation coefficients, absolute values were taken for all correlation coefficients before averaging. The calculation results are shown in Table 1. The results show that the relevance of features in the network decreases with the number of stages in the form of an inverted pyramid, which verifies our expectation that the impact of the classification loss function (Softmax Loss) in the last layer on front layers is gradually attenuated, so the parameter adjustment is probably not optimal.

The joint supervision of Softmax Loss+MFD Loss is used to train the neural network for feature learning, and MFD Loss is placed after multiple network front layers for constraint. Ultimately, the expression of the joint loss is as follows:
\begin{equation}
\begin{split}
L
& = L_{Softmax}+\lambda \sum_{i=1}^s L_{MFD} \\
& = -\frac{1}{b}\sum_{i=1}^b ln{\frac{e^{z_{y_i}}}{\sum_{j}e^{z_j}}}+\lambda \sum_{i=1}^s \frac{\sum_{j}^{d^i}\sum_{k \neq j}^{d^i} (F_{jk}^i)^2}{d^i(d^i-1)}\\
\end{split}
\end{equation}
where $s$ denotes that MFD loss constraints need to be placed behind $s$ front layers, and $\lambda$ is the balance factor.
\begin{figure}[htbp]
\centering
\includegraphics[width=0.45\textwidth]{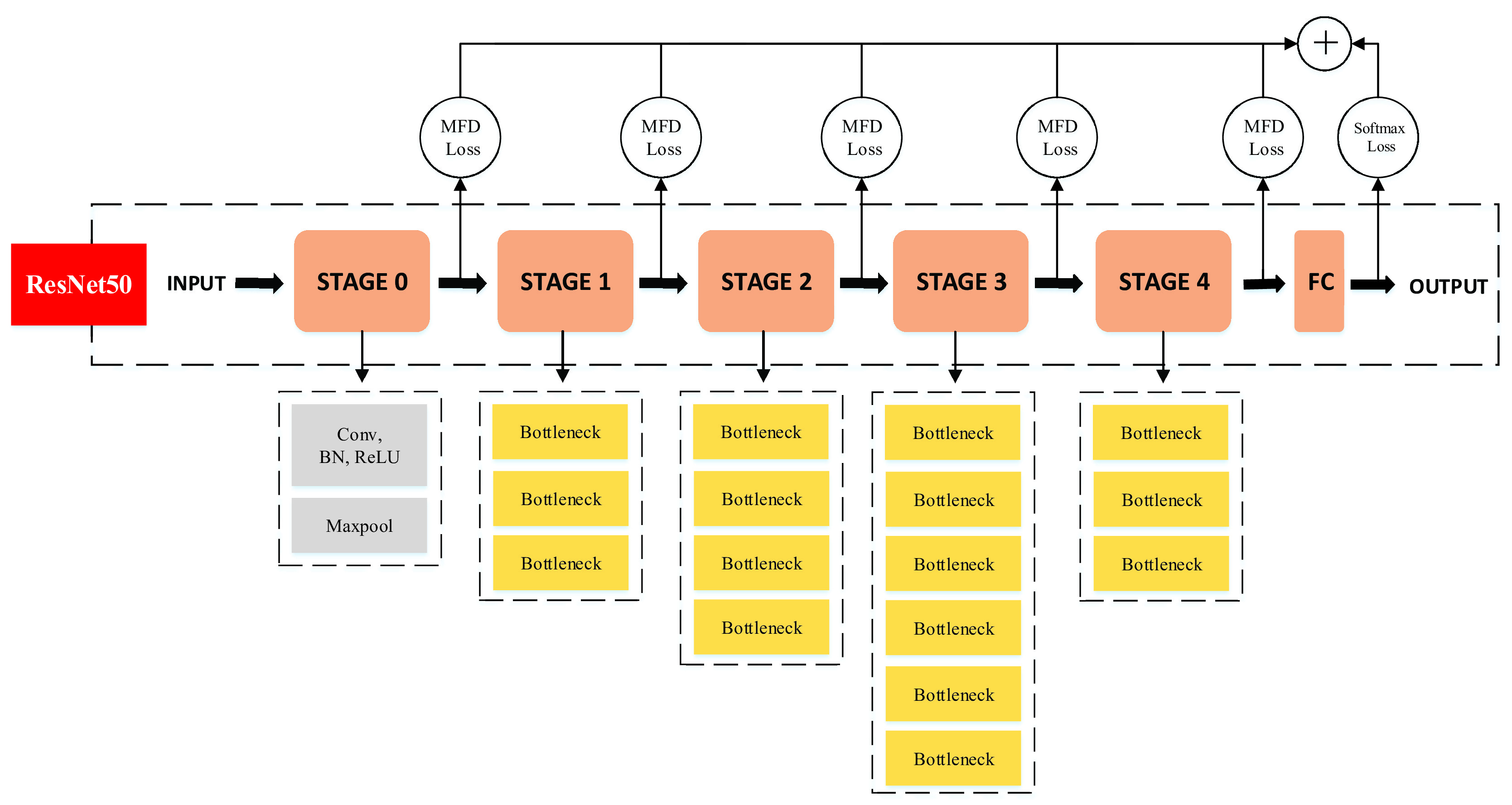}
\caption{ResNet50 network structure and the corresponding MFD loss function and classification loss function.}
\label{fig3}
\end{figure}
\begin{table}[htbp]
\caption{Average values of correlation coefficients between features after Stage 0 to Stage 4 in ResNet50 trained by Softmax Loss}
\begin{center}
\begin{tabular}{cccccc}\hline
Stage No. & 0 & 1 & 2 & 3 & 4\\ \hline
Value & 0.189 & 0.142 & 0.120 & 0.053 & 0.094\\ \hline
\end{tabular}
\label{tab1}
\end{center}
\end{table}
\section{Experiments and Results}
Experiment is implemented using Pytorch1.0~\cite{b16}. The network structures used include ResNet50~\cite{b7}, ResNext50~\cite{b17}, DenseNet121~\cite{b18} and MobileNetV2~\cite{b19}. The datasets used include CIFAR10~\cite{b20}, CIFAR100~\cite{b20}, Tiny ImageNet ~\cite{b21} and FaceScrub~\cite{b22}. In order to make the network structure more suitable for the image sizes of different datasets, some modifications are made to the original ResNet50, ResNext50 and DenseNet121 structures.First, the convolution kernel of the first convolutional layer is changed from the original $7 \times 7$ to $3 \times 3$, and the stride is changed to $1$. Second, the first maxpool layer is also eliminated (except Tiny ImageNet and FaceScrub). All experimental results are the average of three experiments.

\subsection{Experimental Datasets}\label{AA}
CIFAR10 dataset contains $10$ classes of RGB color images, and the images size is $32\times32$. There are $50000$ training images and $10000$ test images in the dataset. CIFAR100 dataset contains $100$ classes of images, the images size is $32\times32$, a total of $50000$ training images and $10000$ test images. FaceScrub dataset contains $100$ classes of images, the images size is $64\times64$, there are $15896$ training images and $3896$ test images in the dataset. Tiny ImageNet dataset contains $200$ classes of images, the images size are $64\times64$, there are a total of $100000$ training images and $10000$ test images. For these datasets, standard data enhancement~\cite{b1} is performed, that is, the training images are filled with $4$ pixels, randomly cropped to the original size, and horizontally flipped with a probability of $0.5$, and the test images are not processed. 

For the above datasets, in the training phase, using the SGD optimizer, the weight decay is $0.0005$, and the momentum is $0.9$. The initial learning rate is $0.1$, and a total of $100$ epochs are trained. At the $30$th, $60$th, and $90$th epoch, the learning rate drops to one-tenth of the original, and the batchsize is set to $128$.
\begin{table}[htbp]
\caption{Comparison of classification accuracy (\%) with different values of balance factor $\lambda$ on ResNet50.}
\begin{center}
\begin{tabular}{ccccccc}\hline
$\lg(\lambda)$ & \makecell{Softmax\\only}  & -2 & -1 & 0 & 1 & 2 \\ \hline
Accuracy & \makecell{76.42} & \makecell{76.91\\(0.49$\uparrow$)} & \makecell{78.27\\(1.85$\uparrow$)} & \textbf{\makecell{79.18\\(2.76$\uparrow$)}} & \makecell{76.86\\(0.44$\uparrow$)} & \makecell{72.20\\(4.22$\downarrow$)}\\ \hline
\end{tabular}
\label{tab2}
\end{center}
\end{table}
\begin{table}[htbp]
\caption{Average values of correlation coefficients between features after Stage 0 to Stage 4 in ResNet50 by joint training with Softmax Loss+MFD Loss.}
\begin{center}
\begin{tabular}{cccccc}\hline
Stage No. & 0 & 1 & 2 & 3 & 4\\ \hline
Value & \makecell{0.007\\(0.182$\downarrow$)} & \makecell{0.015\\(0.127$\downarrow$)} & \makecell{0.034\\(0.086$\downarrow$)} & \makecell{0.031\\(0.022$\downarrow$)} & \makecell{0.013\\(0.081$\downarrow$)}\\ \hline
\end{tabular}
\label{tab3}
\end{center}
\end{table}
\begin{table}
\caption{Comparison of classification accuracy (\%) of Softmax Loss and Softmax Loss+MFD Loss on ResNet50.}
\centering
\setlength{\tabcolsep}{1.5mm}{
\begin{tabular}{ccccc}
\toprule
  \diagbox{Loss}{Dataset} & CIFAR10 & CIFAR100 & Tiny ImageNet & FaceScrub 
  \\
\midrule
 Softmax Loss & 93.25 & 76.42 & 62.10 & 90.56
  \\
  \makecell{Softmax Loss\\+MFD Loss} & \textbf{\makecell{93.98 \\(0.73$\uparrow$)}} & \textbf{\makecell{79.18 \\(2.76$\uparrow$)}} & \textbf{\makecell{64.06 \\(1.96$\uparrow$)}} & \textbf{\makecell{94.63 \\(4.07$\uparrow$)}} 
  \\
\bottomrule
\end{tabular}}
\end{table}
\subsection{Experimental Results}
\subsubsection{Experimental Results on ResNet50}

The neural network is trained with the joint supervision of Softmax Loss and MFD Loss, and MFD Loss is placed after different stages in ResNet50 to constrain the network learning for comparative experiments. Table 2 shows the classification performance on the CIFAR100 dataset with different values of the balance factor $\lambda$. It can be seen from the table that when $\lg(\lambda)$ increases from -2 to 0, the classification accuracy increases steadily and is better than that of Softmax Loss, but the classification accuracy decreases when it continues to increase, and drops at $\lg(\lambda)=1$. Therefore, in the subsequent experiments, we choose $\lg(\lambda)=0$, that is, set $\lambda=1$.

The balance factor $\lambda$ is set to $1$, and MFD Loss is added after Stage $0$ to Stage $4$. After fully training the neural network, the average values of correlation coefficients between features after Stage $0$ to Stage $4$ are calculated, which are shown in Table 3. Table 3 shows that these average values are significantly lower than before, indicating a decrease in the degree of correlation between features, ensuring the validity of features in the network and improving the classification performance.

In order to more intuitively demonstrate the effectiveness of MFD Loss, the comparison of the visualizations of feature maps after single Softmax Loss training and Softmax Loss+MFD Loss training are presented. Due to the larger feature space accumulated by the deeper the network, it is no longer convenient for the human eye to observe and analyze. Therefore, only the feature maps of the first $64$ channels after Stage $0$ and Stage $1$ are given for the comparison. As shown in Fig. 4 and Fig. 5, the introduction of MFD Loss reduces the number of similar feature maps, and the similarity between the non-all-zero-valued feature maps is effectively reduced. Although the all-zero-valued feature maps are increased, the correlation between the all-zero-valued feature maps and any feature maps is computationally zero, so it does not contradict the results in Table 3. Meanwhile, the all-zero-valued feature maps not only have no effect on the classification, but also occupy resources, which means they can be directly discarded or used in subsequent stages of incremental learning. Therefore, we will verify the applicability of MFD Loss in network cropping, incremental learning and so on in subsequent study.

\vspace{-1em} 
\begin{figure}[htbp]
    \begin{minipage}[t]{0.5\linewidth}
        \centering
        \includegraphics[width=\textwidth]{Figure1.jpg}
        \centerline{Softmax Loss}
    \end{minipage}%
    \begin{minipage}[t]{0.5\linewidth}
        \centering
        \includegraphics[width=\textwidth]{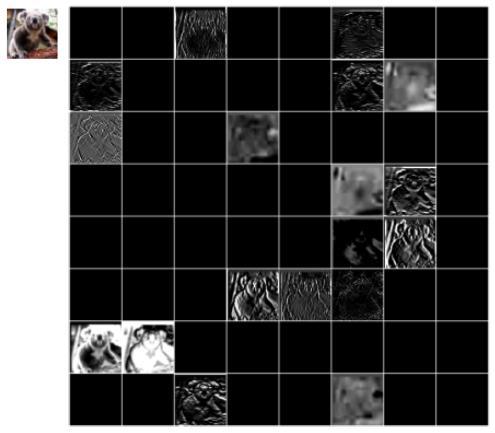}
        \centerline{Softmax Loss+MFD Loss}
    \end{minipage}
    \caption{Visualization of some feature maps generated by Stage $0$ in ResNet50 after full training, where similar feature maps are annotated with boxes of the same color.}
\end{figure}
\vspace{-1em} 
The experiments on CIFAR10, FaceScrub, and Tiny ImageNet are also performed. The comparison experimental results of Softmax Loss and Softmax Loss+MFD Loss are shown in Table 4, which shows that the classification performance of the joint supervision of Softmax Loss+MFD Loss is higher than that of Softmax Loss alone on multiple datasets.

\subsubsection{Experimental Results on Other CNNs}
Based on the different feature map size, ResNext50, DenseNet121 and MobileNetV2 are all divided into $5$ stages. Similarly, MFD Loss constraints is added in each stage, and the comparison results between Softmax Loss and Softmax Loss+MFD Loss are shown in Table 5. The results show that the joint supervision of Softmax Loss + MFD Loss has higher classification performance on ResNeXt50, DenseNet121 and MobileNetV2, which indicates that the addition of MFD Loss can effectively improve the classification accuracy on traditional CNN.
\vspace{-1em} 
\begin{figure}[htbp]
    \begin{minipage}[t]{0.5\linewidth}
        \centering
        \includegraphics[width=\textwidth]{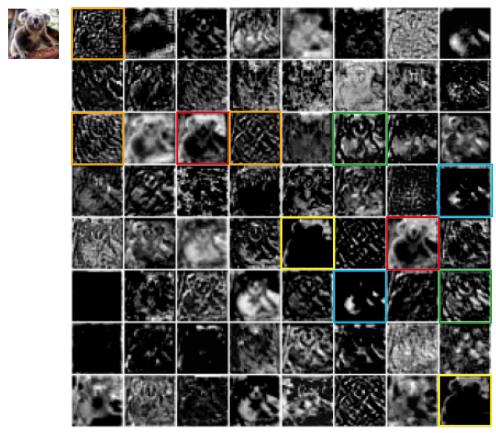}
        \centerline{Softmax Loss}
    \end{minipage}%
    \begin{minipage}[t]{0.5\linewidth}
        \centering
        \includegraphics[width=\textwidth]{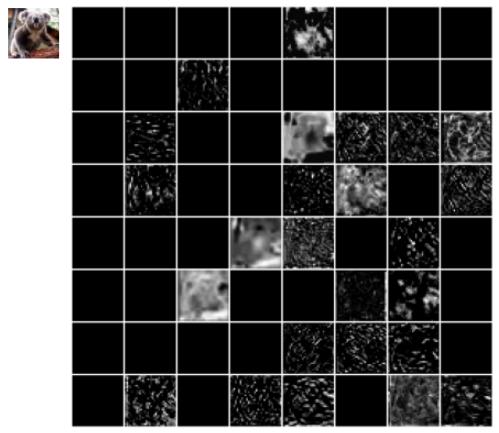}
        \centerline{Softmax Loss+MFD Loss}
    \end{minipage}
    \caption{Visualization of some feature maps generated by Stage $1$ in ResNet50 after full training, where similar feature maps are annotated with boxes of the same color.}
\end{figure}
\vspace{-1em} 
\begin{table}[H]
\caption{Comparison of classification accuracy (\%) of Softmax Loss and Softmax Loss+MFD Loss on different CNNs.}
\centering
\small
\setlength{\tabcolsep}{2mm}{
\begin{tabular}{cccc}
\toprule
 Network & Dataset & \makecell{Accuracy\\(Softmax Loss)} & \makecell{Accuracy\\(Softmax Loss\\+MFD Loss)} \\
\midrule
\multirow{4}*{ResNeXt50} & CIFAR10 & 94.05 & \textbf{94.61 (0.56$\uparrow$)} \\
~ & CIFAR100 & 78.57 & \textbf{79.71  (1.14$\uparrow$)} \\
~ & Tiny ImageNet &64.23 & \textbf{65.63  (1.40$\uparrow$)} \\
~ & FaceScrub & 93.36 & \textbf{95.09  (1.73$\uparrow$)} \\
\midrule
  \multirow{4}*{DenseNet121} & CIFAR10 & 95.01 &  \textbf{95.57   (0.56$\uparrow$)} 
  \\
  ~ & CIFAR100 & 79.91 & \textbf{80.54   (0.63$\uparrow$)} 
  \\
  ~ & Tiny ImageNet & 64.73 & \textbf{65.59  (0.86$\uparrow$)} 
  \\
  ~ & FaceScrub & 95.61 & \textbf{96.27  (0.66$\uparrow$)} 
  \\
\midrule
  \multirow{4}*{MobileNetV2} & CIFAR10 & 84.31 &  \textbf{86.63  (2.32$\uparrow$)} 
  \\
  ~ & CIFAR100 & 52.41 & \textbf{54.99  (2.58$\uparrow$)} 
  \\
  ~ & Tiny ImageNet & 46.34 & \textbf{49.14  (2.80$\uparrow$)} 
  \\
  ~ & FaceScrub & 87.84 & \textbf{90.32 (2.48$\uparrow$)} 
  \\
\bottomrule
\end{tabular}}
\end{table}

\subsubsection{Effect of Combining with Loss Functions Dedicated to Face Recognition}
To verify the generalizability of MFD Loss, Center Loss, AM-Softmax ($m=0.35$), and ArcFace ($m=0.5$) are combined with MFD Loss to conduct comparison experiments on ResNet50 network and face recognition dataset FaceScrub, respectively. The experimental results are shown in Table 6. It can be seen that for face recognition task, adding MFD Loss consistently helps to improve the classification performance of the network compared to training with different and single loss functions, which indicates that MFD Loss has good generalizability.
\begin{table}
\caption{Comparison of classification accuracy (\%) between different loss functions dedicated to face recognition before and after combining with MFD Loss on ResNet50 network and FaceScrub dataset.}\label{tab6}
\centering
\setlength{\tabcolsep}{4mm}{
\begin{tabular}{cc}
\toprule
Loss Function & Accuracy\\
\midrule
Center Loss & 92.82 \\
Center Loss+MFD Loss & \textbf{94.11(1.29$\uparrow$)}\\
\midrule
AM-Softmax($m=0.35$) & 93.60 \\
AM-Softmax($m=0.35$)+MFD Loss & \textbf{94.47(0.87$\uparrow$)}\\
\midrule
ArcFace($m=0.5$) & 93.93 \\
ArcFace($m=0.5$)+MFD Loss & \textbf{94.89(0.96$\uparrow$)}\\
\bottomrule
\end{tabular}}
\end{table}
\begin{figure}[htbp]
\centering
\includegraphics[width=0.35\textwidth]{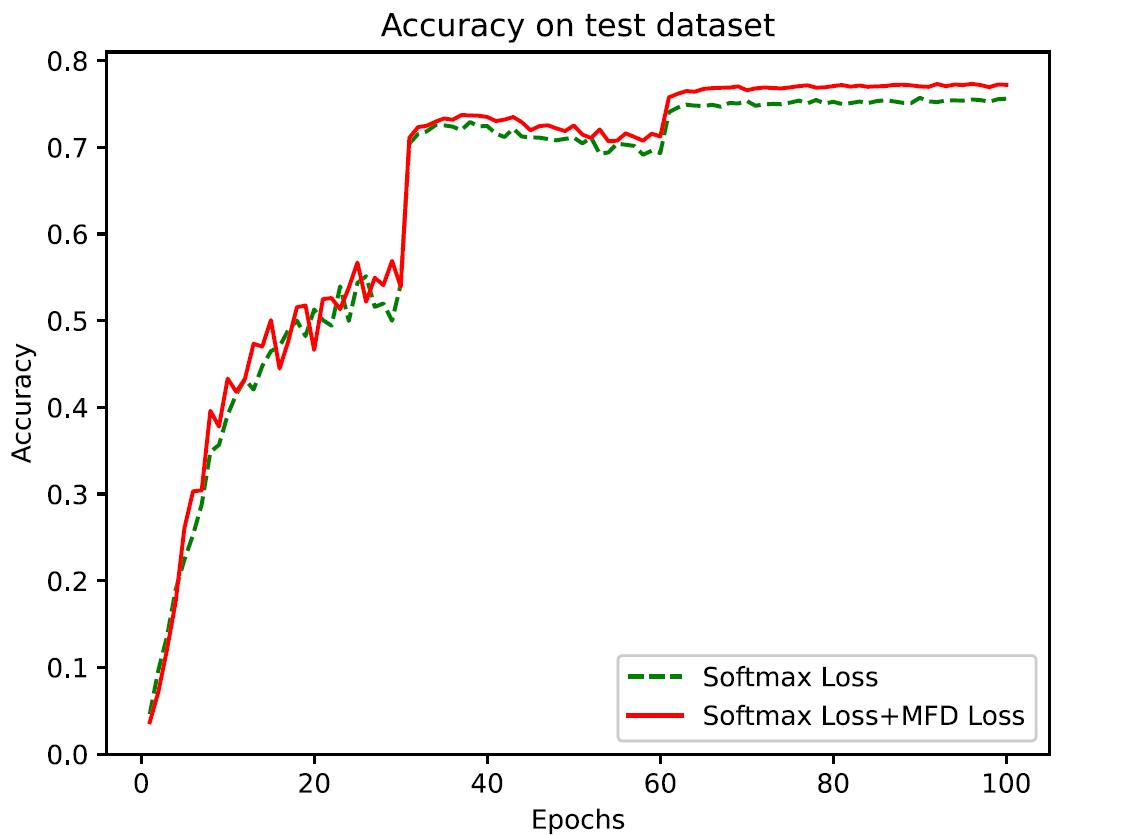}
\caption{Variation of classification accuracy of various loss functions with epoch during training.}
\label{fig6}
\end{figure}
\begin{table*}
\centering
\caption{Comparison of one epoch’s training consumption time and GPU memory usage of various loss functions on different datasets}
\begin{tabular}{c|c|c|c|c} 
\toprule 
\multirow{2}*{Dataset} & \multicolumn{2}{c}{One epoch’s training time consumption(s)} & \multicolumn{2}{c}{GPU memory usage(MiB)}\\ 

\cline{2-3}
\cline{4-5}\\

& Softmax Loss & Softmax Loss +MFD Loss & Softmax Loss & Softmax Loss +MFD Loss \\

\midrule 
CIFAR10 & 69 & 105(52.17\%$\uparrow$) & 11322 & 11316(0.05\%$\downarrow$) \\ 
CIFAR100 & 69 & 105(52.17\%$\uparrow$) & 11304 & 11046(2.28\%$\downarrow$) \\
Tiny ImageNet & 146 & 237(62.33\%$\uparrow$) & 11664 & 11666(0.02\%$\uparrow$) \\
FaceScrub & 23 & 38(65.22\%$\uparrow$) & 12146 & 12570(3.49\%$\uparrow$) \\
\bottomrule 
\end{tabular}
\end{table*}
\subsubsection{Comparison of other performance indexes}
On ResNet50, the comparative experiments on convergence, training speed and GPU memory usage are performed. Figure 6 shows the comparison of the convergence speed between Softmax Loss and Softmax Loss+MFD Loss during the training of the network on the CIFAR100 dataset. It can be seen that there is little difference in the convergence rates between the two, but the joint supervision of Softmax Loss+MFD Loss obviously has a higher classification accuracy in the middle and later stages of training. Table 7 shows the comparison results of one epoch’s training consumption time and GPU memory usage on different datasets. The addition of MFD Loss increases the back propagation time of the network, thus reducing the training speed of the network to some extent, but with little additional overhead in terms of GPU memory usage.

\section{Conclusion}
For CNN, this article proposes a new loss function—MFD Loss, which constrains multiple front layers of the network. MFD Loss is placed after several front stages of the network to supervise learning, which solves the problem that the impact of traditional loss functions on front layers is gradually reduced due to the large depth of network. Through the experiment, we find that feature mappings have significant redundancy. MFD Loss constrains the correlation between features to improve the validity of these feature mappings, and the joint supervised learning of MFD Loss and Softmax Loss further improves the classification performance. The experimental results show that, compared with single Softmax Loss supervised learning, Softmax Loss+MFD Loss can achieve better performance on image classification tasks. Meanwhile, the good generalizability of MFD Loss is verified on other typical classification loss functions. In future work, we will study the applicability of MFD Loss on Transformer networks, and practice the applicability of MFD Loss in model reduction and incremental learning.

\vspace{12pt}
\color{red}

\end{document}